\documentclass[11pt,a4paper]{article}
\usepackage[hyperref]{emnlp-ijcnlp-2019}
\usepackage{times}
\usepackage{latexsym}
\usepackage{pifont}

\usepackage{etex}
\usepackage{booktabs}
\usepackage{graphicx}
\usepackage{subcaption}
\usepackage[export]{adjustbox}
\usepackage{microtype}

\usepackage{url}
\usepackage{xspace}
\usepackage{color}
\usepackage{xcolor}
\usepackage{wrapfig}
\usepackage{algorithm}
\usepackage[noend]{algpseudocode}
\usepackage[T2A,T1]{fontenc}
\usepackage[utf8]{inputenc}
\usepackage[russian,english]{babel}
\usepackage{arydshln}
\usepackage{array}
\aclfinalcopy %

\usepackage{balance}

\newcommand\tagger{{\sc LaserTagger}\xspace}
\newcommand\taggerAR{{\sc LaserTagger}\textsubscript{AR}\xspace}
\newcommand\taggerFF{{\sc LaserTagger}\textsubscript{FF}\xspace}
\newcommand\seqtoseqbert{{\sc seq2seq}\textsubscript{BERT}\xspace}

\newcommand{\tag}[1]{\texttt{#1}\xspace}
\newcommand\ssplit{$\langle$::::$\rangle$\xspace}
\newcommand{\tagadd}[2]{{}\textsuperscript{#2}\texttt{#1}\xspace}
\newtheorem{pproblem}{Problem}
\newtheorem{theorem}{Theorem}

\definecolor{darkgreen}{rgb}{0, 0.5, 0}

\title{Encode, Tag, Realize: High-Precision Text Editing}

\author{Eric Malmi \\
  Google Research \\
  {\tt emalmi@google.com} \\\And
  Sebastian Krause \\
  Google Research \\
  {\tt bastik@google.com} \\\And
  Sascha Rothe \\
  Google Research \\
  {\tt rothe@google.com} \\\AND
  Daniil Mirylenka \\
  Google Research \\
  {\tt dmirylenka@google.com} \\\And
  Aliaksei Severyn \\
  Google Research \\
  {\tt severyn@google.com}}

\date{}

\begin{document}
\maketitle
\begin{abstract}
We propose \tagger---a sequence tagging approach that casts text generation as a text editing task.
Target texts are reconstructed from the inputs using three main edit operations: \textit{keeping} a token, \textit{deleting} it, and \textit{adding} a phrase before the token.
To predict the edit operations, we propose a novel model, which combines a BERT encoder with an autoregressive Transformer decoder.
This approach is evaluated on English text on four tasks:
sentence fusion, sentence splitting, abstractive summarization, and grammar correction.
\tagger achieves new state-of-the-art results on three of these tasks,
performs comparably to a set of strong seq2seq baselines with a large number of training examples, and outperforms them when the number of examples is limited.
Furthermore, we show that at inference time tagging can be more than two orders of magnitude faster than comparable seq2seq models, making it more attractive for running in a live environment.
\end{abstract}

\section{Introduction}

Neural sequence-to-sequence (seq2seq) models provide a powerful framework for learning to translate source texts into target texts.
Since their first application to machine translation (MT)~\cite{sutskever} they have become the de facto approach for virtually every text generation task, including summarization \citep{Tan2017-tt}, image captioning \citep{captioning}, text style transfer~\cite{rao2018dear,nikolov2018large,jin2019unsupervised}, and grammatical error correction \citep{chollampatt-ng-2018,grundkiewicz2019neural}.

We observe that in some text generation tasks, such as the recently introduced \textit{sentence splitting} and \textit{sentence fusion} tasks, output texts highly overlap with inputs.
In this setting, learning a seq2seq model to generate the output text from scratch seems intuitively wasteful.
Copy mechanisms
~\cite{gu2016incorporating,See2017-ue}
allow for choosing between copying source tokens and generating arbitrary tokens, but
although such hybrid models help with out-of-vocabulary words, they still require large training sets as they depend on output vocabularies as large as those used by the standard seq2seq approaches.

In contrast, we propose learning a text editing model
that applies a set of edit operations on the input sequence to reconstruct the output.  
We show that it is often enough to use a relatively small set of output tags representing text deletion, rephrasing and word reordering to be able to reproduce a large percentage of the targets in the training data.
This results in a learning problem
with a much smaller vocabulary size,
and the output length fixed to the number of words in the source text.
This, in turn, greatly reduces the number training examples required to 
train accurate models, which is particularly important in applications where only a small number of human-labeled data is available.

\begin{figure}[tb]
\includegraphics[trim=0 20 20 5,width=\linewidth]{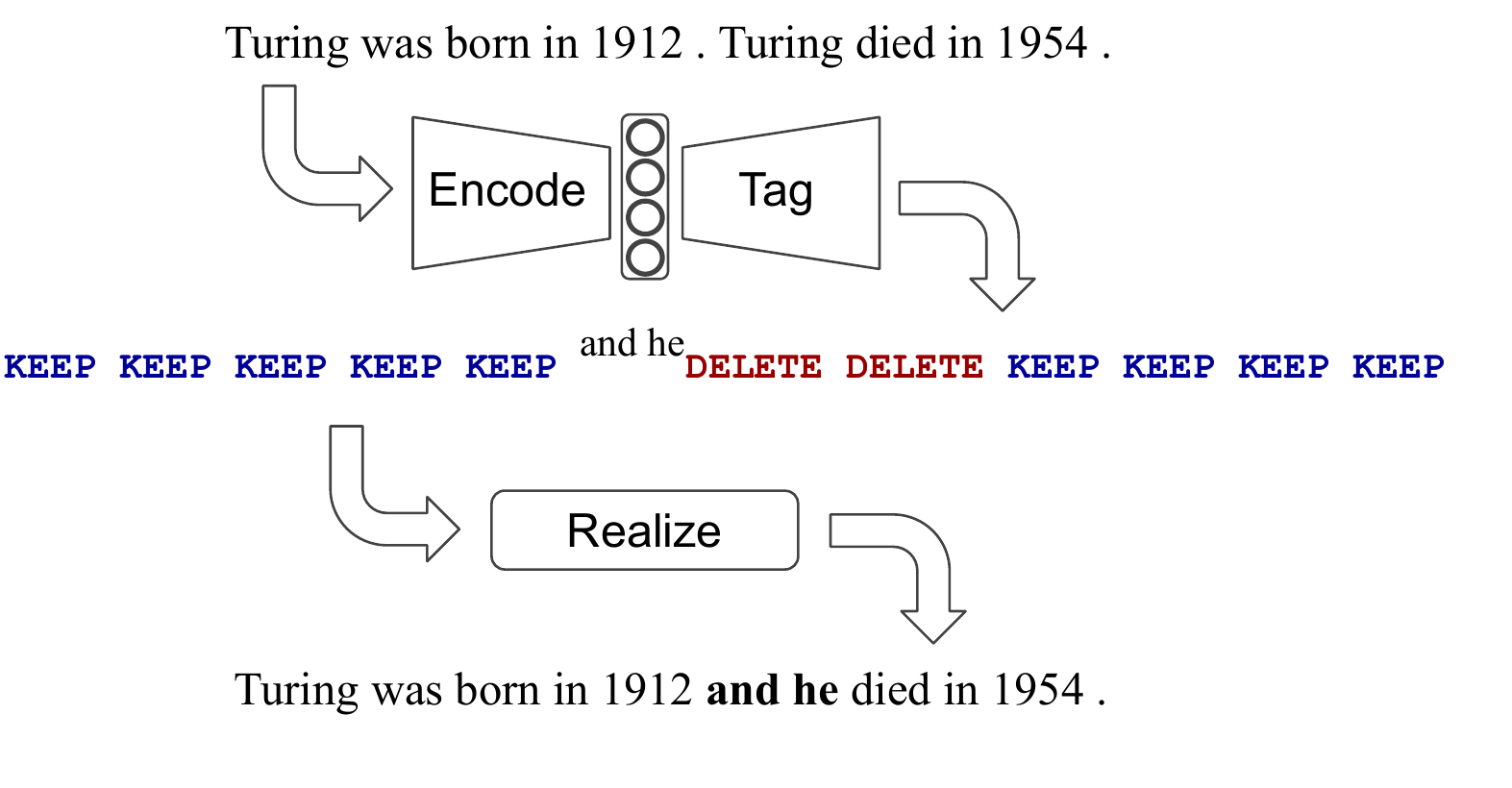}
\caption{\tagger applied to sentence fusion.}
\label{fig:tagger_overview}
\end{figure}

Our tagging approach, \tagger, consists of three steps (Fig.~\ref{fig:tagger_overview}): (i) \textit{Encode} builds a representation of the input sequence, (ii) \textit{Tag} assigns edit tags from a pre-computed output vocabulary to the input tokens, and (iii) \textit{Realize} applies a simple set of rules to convert tags into the output text tokens.

An experimental evaluation of \tagger on four different text generation tasks shows that it yields comparable results to seq2seq models when we have tens of thousands of training examples and clearly outperforms them when the number of examples is smaller.

Our contributions are the following:

\textbf{1)} We demonstrate that many text generation tasks with overlapping inputs and outputs can be effectively treated as text editing tasks.

\textbf{2)} We propose \tagger---a sequence tagging-based model for text editing, together with a method for generating the tag vocabulary from the training data.

\textbf{3)} We describe two versions of the tagging model: (i)~\taggerFF{}---a tagger based on BERT ~\cite{devlin2018bert} and (ii)~\taggerAR{}---a novel tagging model combining the BERT encoder with an autoregressive Transformer decoder, which further improves the results over the BERT tagger.

\textbf{4)} We evaluate \tagger against strong seq2seq baseline models based on the BERT architecture. Our baseline models outperform previously reported state-of-the-art results on two tasks.

\textbf{5)} We demonstrate that a)~\taggerAR{} achieves state-of-the-art or comparable results on 3 out of 4 examined tasks, b)~\taggerFF{} is up to \textit{100x faster at inference time} with performance comparable to the state-of-the-art seq2seq models. Furthermore, both models: c) require much \textit{less training data} compared to the seq2seq models, d) are more \textit{controllable and interpretable} than seq2seq models due to the small vocabulary of edit operations, e) are \textit{less prone} to typical seq2seq model errors, such as hallucination.

The code will be available at: \\ \url{lasertagger.page.link/code}

\section{Related Work}

Recent work discusses some of the difficulties of learning neural decoders for text generation \citep{wiseman2018learning,prabhakaran-etal-2018-detecting}. Conventional seq2seq approaches require large amounts of training data, are hard to control and to constrain to desirable outputs.
At the same time, many NLP tasks that appear to be full-fledged text generation tasks are natural testbeds for simpler methods. %
In this section we briefly review some of these tasks.

\textbf{Text Simplification} is a paraphrasing task that is known to benefit from modeling edit operations. A simple instance of this type are sentence compression systems that apply a drop operation at the token/phrase level \citep{Filippova2008-qk,Filippova2015-xd}, while more intricate systems also apply splitting, reordering, and lexical substitution \citep{Zhu2010-mv}. Simplification has also been attempted with systems developed for phrase-based MT \citep{Xu2016-mu}, as well as with neural encoder-decoder models \citep{Zhang2017-th}.

Independent of this work, \citet{dong2019editnts} recently proposed a text-editing model, similar to ours, for text simplification. The main differences to our work are: ($i$)~They introduce an interpreter module which acts as a language model for the so-far-realized text, and ($ii$)~they generate added tokens one-by-one from a full vocabulary rather than from an optimized set of frequently added phrases. The latter allows their model to generate more diverse output, but it may negatively effect the inference time, precision, and the data efficiency of their model.
Another recent model similar to ours is called Levenshtein Transformer \citet{gu2019levenshtein}, which does text editing by performing a sequence of deletion and insertion actions.

\textbf{Single-document summarization} is a task that requires systems to shorten texts in a meaning-preserving way. It has been approached with deletion-based methods on the token level \citep{Filippova2015-xd} and the sentence level \citep{Narayan2018-nn,Liu2019-ut}. Other papers have used neural encoder-decoder methods \citep{Tan2017-tt,Rush2015-qf,Paulus2017-dx} to do \emph{abstractive} summarization, which allows edits beyond mere deletion. This can be motivated by the work of \citet{Jing2000-hj}, who identified a small number of fundamental high-level editing operations that are useful for producing summaries (reduction, combination, syntactic transformation, lexical paraphrasing, generalization/specification, and reordering).
\citet{See2017-ue} extended a neural encoder-decoder model with a copy mechanism to allow the model to more easily reproduce input tokens during generation.

Out of available summarization datasets \citep{Dernoncourt2018-op}, we find the one by 
\citet{Toutanova2016-rs} particularly interesting because (1) it specifically targets abstractive summarization systems, (2) the lengths of texts in this dataset (short paragraphs) seem well-suited for text editing, and (3) an analysis showed that the dataset covers many different summarization operations.

In \textbf{Grammatical Error Correction} \citep{ng-etal-2013-conll,ng-etal-2014-conll} a system is presented with input texts written usually by a language learner, and is tasked with detecting and fixing grammatical (and other) mistakes. Approaches to this task often incorporate task-specific knowledge, e.g., by designing classifiers for specific error types \citep{KnightC94,rozovskaya-etal-2014-illinois} that can be trained without manually labeled data,
or by adapting statistical machine-translation methods \citep{junczys-dowmunt-grundkiewicz-2014-amu}. Methods for the sub-problem of error \emph{detection} are similar in spirit to sentence compression systems, in that they are implemented as word-based neural sequence labelers \citep{rei-2017,rei-etal-2017}. Neural encoder-decoder methods are also commonly applied to the error correction task \citep{ge-etal-2018-fluency,chollampatt-ng-2018,zhaoAbs1903}, but suffer from a lack of training data, which is why task-specific tricks need to be applied \citep{kasewa-etal-2018-wronging,junczys-dowmunt-etal-2018-approaching}.

\begin{figure*}[t]
\setlength{\tabcolsep}{4pt}
\centering
\resizebox{\textwidth}{!}{
\begin{tabular}{llcccccccccc}
\textbf{Source:} & Dylan & won & Nobel & prize & . & Dylan & is & an & American & musician & . \\
\textbf{Tags:} & \tag{\color{red}DELETE} & \tag{\color{blue}KEEP} & \tag{\color{blue}KEEP} & \tag{\color{blue}KEEP} & \tag{SWAP} & \tag{\color{blue}KEEP} & \tagadd{\color{red}DELETE}{\textit{comma}} & \tag{\color{blue}KEEP} & \tag{\color{blue}KEEP} & \tag{\color{blue}KEEP} & \tagadd{\color{red}DELETE}{\textit{comma}}  \\
\textbf{Realization:} & \multicolumn{11}{l}{Dylan , an American musician , won Nobel prize .} \\
\end{tabular}
}
\caption{An example sentence fusion obtained by tagging using the \tag{SWAP} tag, which swaps the order of the two source sentences.}\label{fig:swap_example}
\end{figure*}

\section{Text Editing as a Tagging Problem}

Our approach to text editing is to cast it into a tagging problem. Here we describe its main components: (1) the tagging operations, (2) how to convert plain-text training targets into a tagging format, as well as (3) the realization step to convert tags into the final output text.

\subsection{Tagging Operations} \label{sec:operations}
Our tagger assigns a tag to each input token. 
A tag is composed of two parts: a \textit{base tag} and an \textit{added phrase}.
The base tag is either \tag{KEEP} or \tag{DELETE}, which indicates whether to retain the token in the output.
The added phrase $P$, which can be empty, enforces that $P$ is added before the corresponding token.
$P$ belongs to a vocabulary $V$ that defines a set of words and phrases that can be inserted into the input sequence to transform it into the output.

The combination of the base tag $B$ and the added phrase $P$ is treated as a single tag and denoted by \tagadd{$B$}{$P$}. The total number of unique tags is equal to the number of base tags times the size of the phrase vocabulary, hence there are $\approx{}2|V|$ unique tags. 

Additional task-specific tags can be employed too.
For sentence fusion (Section~\ref{sec:fusion}), the input consists of two sentences, which sometimes need to be swapped.
Therefore, we introduce a custom tag, \tag{SWAP}, which can only be applied to the last period of the first sentence (see Fig.~\ref{fig:swap_example}).
This tag instructs the Realize step to swap the order of the input sentences before realizing the rest of the tags.

For other tasks, different supplementary tags may be useful. E.g., to allow for replacing entity mentions with the appropriate pronouns, we could introduce a \tag{PRONOMINALIZE} tag. Given an access to a knowledge base that includes entity gender information, we could then look up the correct pronoun during the realization step, instead of having to rely on the model predicting the correct tag (\tagadd{DELETE}{she}, \tagadd{DELETE}{he}, \tagadd{DELETE}{they}, etc.).

\subsection{Optimizing Phrase Vocabulary} \label{sec:vocab}
The phrase vocabulary consists of phrases that can be added between the source words.
On the one hand, we wish to minimize the number of phrases to keep the output tag vocabulary small. On the other hand, we would like to maximize the percentage of target texts that can be reconstructed from the source using the available tagging operations. %
This leads to the following combinatorial optimization problem.

\begin{pproblem} \label{problem:vocab}
Given a collection of phrase sets $A_1, A_2, \ldots A_m$, where $A_i \subseteq P$ and $P$ is the set of all candidate phrases, select a phrase vocabulary $V \subset P$ of at most $\ell$ phrases (i.e. $|V| \leq \ell$) so that the number of covered phrase sets is maximized. A phrase set $A_i$ is covered if and only if $A_i \subseteq V$.
\end{pproblem}

This problem is closely related to the minimum $k$-union problem which is NP-hard \cite{vinterbo2002note}. The latter problem asks for a set of $k$ phrase sets such that the cardinality of their union is the minimum. If we were able to solve Problem~\ref{problem:vocab} in polynomial time, we could solve also the minimum $k$-union problem in polynomial time simply by finding the smallest phrase vocabulary size $\ell$ such that the number of covered phrase sets is at least $k$. This reduction from the minimum $k$-union problem gives us the following result:

\begin{theorem}
Problem~\ref{problem:vocab} is NP-hard.
\end{theorem}

To identify candidate phrases to be included in the vocabulary, we first align each source text $s$ from the training data with its target text $t$. This is achieved by computing the \textit{longest common subsequence} (LCS) between the two word sequences, which can be done using dynamic programming in time $\mathcal{O}(|s| \times |t|)$.
The n-grams in the target text that are not part of the LCS are the phrases that would need to be included in the phrase vocabulary to be able to construct $t$ from $s$. 

In practice, the phrase vocabulary is expected to consist of phrases that are frequently added to the target. Thus we adopt the following simple approach to construct the phrase vocabulary: sort the phrases by the number of phrase sets in which they occur and pick $\ell$ most frequent phrases. This was found to produce meaningful phrase vocabularies based on manual inspection as shown in Section~\ref{sec:experiments}. E.g., the top phrases for sentence fusions include many discourse connectives.

We also considered a greedy approach that constructs the vocabulary one phrase at a time, always selecting the phrase that has the largest incremental coverage. This approach is not, however, ideal for our use case, since some frequent phrases, such as ``\texttt{(}'' and ``\texttt{)}'', are strongly coupled. Selecting ``\texttt{(}'' alone has close to zero incremental coverage, but together with ``\texttt{)}'', they can cover many examples.

\subsection{Converting Training Targets into Tags}
\label{sec:converting-training-targets-into-tags}

Once the phrase vocabulary is determined, we can convert the target texts in our training data into tag sequences. Given the phrase vocabulary, we do not need to compute the LCS, but can leverage a more efficient approach, which iterates over words in the input and greedily attempts to match them (1)~against the words in the target, and in case there is no match, (2)~against the phrases in the vocabulary $V$. This can be done in $\mathcal{O}(|s| \times n_p)$ time, where $n_p$ is the length of the longest phrase in $V$, as shown in Algorithm~\ref{alg:conversion}.

\begin{algorithm}[htb]
  \small
  \begin{algorithmic}[1]
      \Statex {\bf Input:} Source text $s=[s(1), \ldots, s({n_s})]$, target text $t=[t(1), \ldots, t({n_t})]$, phrase vocabulary $V$, and the maximum added phrase length $n_p$.
      \Statex {\bf Output:} Tag sequence $x$ of length $n_s$ or of length 0 if conversion is not possible. 
      \State$x(i) = \tag{DELETE},\quad \forall i=1, \ldots, n_s$  \Comment{Initialize tags.}
      \State $i_s = 1$ \Comment{Current source word index.}
      \State $i_t = 1$ \Comment{Current target word index.}
      \While{$i_t \leq n_t$}
        \If{$i_s > n_s$}
          \State {\bf return} $[]$ \Comment{Conversion infeasibile.} %
        \EndIf
        \If{$s(i_s) == t(i_t)$}
          \State $x(i_s) =$ \tag{KEEP}
          \State $i_t = i_t + 1$
        \Else
          \State $p = []$ \Comment{Added phrase (word sequence).}
          \State match\_found $= 0$
          \For{$j = 1, \ldots, n_p$}
            \State $p$.append($t(i_t + j - 1$))
            \If{$s(i_s) == t(i_t + j)$ {\bf and} $p \in V$}
              \State match\_found $= 1$
              \State {\bf break}
            \EndIf
          \EndFor
          \If{match\_found}
            \State $x(i_s) = $ \tagadd{KEEP}{$p$}
            \State $i_t = i_t + |p| + 1$
          \EndIf
        \EndIf
        \State $i_s = i_s + 1$
      \EndWhile
      \State {\bf return} $x$ \Comment{Target has been consumed, so return tags.}
  \end{algorithmic}
  \caption{Converting a target string to tags.} \label{alg:conversion}
\end{algorithm}

The training targets that would require adding a phrase that is not in our vocabulary $V$, will not get converted into a tag sequence but are filtered out. While making the training dataset smaller, this may effectively also filter out low-quality targets. The percentage of converted examples for different datasets is reported later in Section~\ref{sec:experiments}. Note that even when the target cannot be reconstructed from the inputs using our output tag vocabulary, our approach might still produce reasonable outputs with the available phrases. E.g., a target may require the use of the infrequent ``\texttt{;}'' token, which is not in our vocabulary, but a model could instead choose to predict a more common ``\texttt{,}'' token.

\subsection{Realization}
After obtaining a predicted tag sequence, we convert it to text (``realization'' step). While classic works on text generation make a distinction between \textit{planning} and \textit{realization}, end-to-end neural approaches typically ignore this distinction, with the exception of few works~\cite{moryossef2019step,puduppully2018data}.

For the basic tagging operations of keeping, deleting, and adding, realization is a straightforward process.
Additionally, we adjust capitalization at sentence boundaries.
Realization becomes more involved if we introduce special tags, such as \tag{PRONOMINALIZE} mentioned in Section~\ref{sec:operations}. For this tag, we would need to look up the gender of the tagged entity from a knowledge base.
Having a separate realization step is beneficial, since we can decide to pronominalize only when confident about the appropriate pronoun and can, otherwise, leave the entity mention untouched.

Another advantage of having a separate realization step is
that specific loss patterns can be addressed by adding specialized realization rules.
For instance, one could have a rule that when applying tag \tagadd{DELETE}{his} to an entity mention followed by \textit{'s}, the realizer must always \tag{DELETE} the possessive \textit{'s} regardless of its predicted tag.

\section{Tagging Model Architecture}

\begin{figure}[tb]
\includegraphics[width=0.9\linewidth]{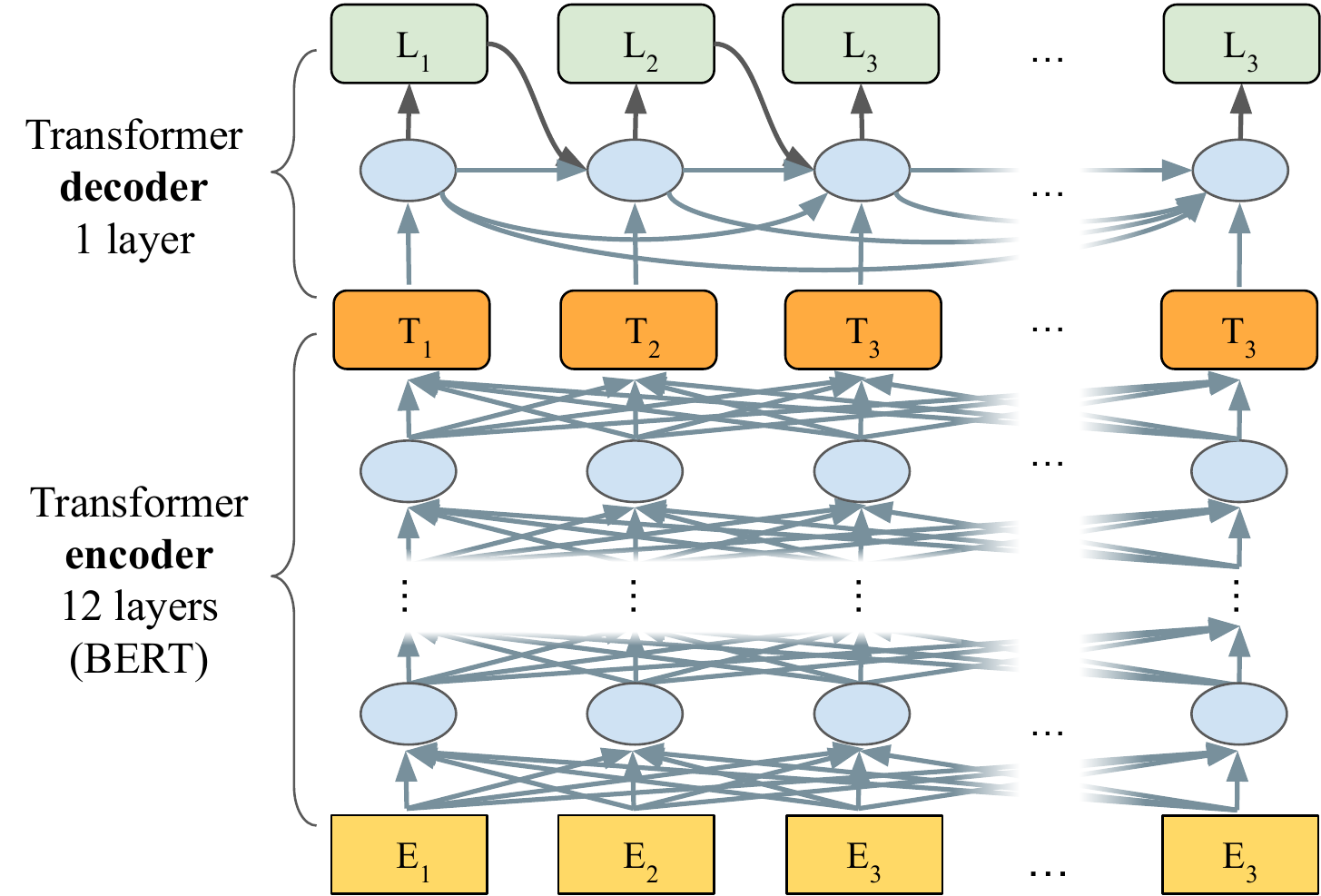}
\caption{The architecture of \taggerAR.}
\label{fig:tagger}
\end{figure}

Our tagger is composed of two components: an encoder, which generates activation vectors for each element in the input sequence, and a decoder, which converts encoder activations into tag labels. 

\textbf{Encoder.}
We choose the BERT Transformer model~\cite{devlin2018bert} as our encoder, as it demonstrated state-of-the-art results on a number of sentence encoding tasks. We use the \texttt{BERT-base} architecture, which consists of 12 self-attention layers.
We refer the reader to~\cite{devlin2018bert} for a detailed description of the model architecture and its input representation. We initialize the encoder with a publicly available checkpoint of the pretrained case-sensitive BERT-base model.\footnote{\scriptsize\url{github.com/google-research/bert}}

\textbf{Decoder.}
In the original BERT paper a simple decoding mechanism is used for sequence tagging: 
the output tags are generated in a single feed-forward pass by applying an \texttt{argmax} over the encoder logits.
In this way, each output tag is predicted independently, without modelling the dependencies between the tags in the sequence.
Such a simple decoder demonstrated state-of-the-art results on the Named Entity Recognition task, when applied on top of the BERT encoder.

To better model the dependencies between the output tag labels, we propose a more powerful autoregressive decoder. Specifically, we run a single-layer Transformer decoder on top of the BERT encoder (see Fig.~\ref{fig:tagger}). At each step, the decoder is consuming the embedding of the previously predicted label and the activations from the encoder. 

There are several ways in which the decoder can communicate with the encoder: (i) through a full attention over the sequence of encoder activations (similar to conventional seq2seq architectures); and (ii) by directly consuming the encoder activation at the current step. In our preliminary experiments, we found the latter option to perform better and converge faster, as it does not require learning additional encoder-decoder attention weights.

We experiment with both decoder variants (\textit{feedforward} and \textit{autoregressive}) and find that the autoregressive decoder outperforms the previously used \textit{feedforward} decoder.
In the rest of this paper, the tagging model with an \textit{autoregressive} decoder is referred to as \taggerAR and the model with \textit{feedforward} decoder as \taggerFF.

\section{Experiments} \label{sec:experiments}

We evaluate our method by conducting experiments on four different text editing tasks: Sentence Fusion, Split and Rephrase, Abstractive Summarization, and Grammatical Error Correction.

\textbf{Baselines.}
In addition to reporting previously published results for each task, we also train a set of strong baselines based on Transformer where both the encoder and decoder replicate the BERT-base architecture~\cite{devlin2018bert}. To have a fair comparison, similar to how we initialize a tagger encoder with a pretrained BERT checkpoint, we use the same initialization for the Transformer encoder. This produces a very strong seq2seq baseline (\seqtoseqbert), which already results in new state-of-the-art metrics on two out of four tasks. 

\subsection{Sentence Fusion}
\label{sec:fusion}

Sentence Fusion is the problem of fusing sentences into a single coherent sentence.

\textbf{Data.}
We use the ``balanced Wikipedia'' portion of \citet{geva2019discofuse}'s DiscoFuse dataset for our experiments (henceforth DfWiki). Out of the 4.5M fusion examples in the dataset, 10.5\% require reordering of the input. To cope with this, we introduce the \tag{SWAP} tag, which enables the model to flip the order of two input sentences. We construct the phrase vocabulary as described in Sec.~\ref{sec:vocab} using the validation set of 46K examples. The top 15 phrases are shown in the first column of Table~\ref{tab:phrases}.

\textbf{Evaluation metrics.}
Following \citet{geva2019discofuse}, we use two evaluation metrics: \textit{Exact score}, which is the percentage of exactly correctly predicted fusions, and \textit{SARI}~\cite{xu2016optimizing}, which computes the average F1 scores of the added, kept, and deleted n-grams.\footnote{We use the implementation available at
\url{git.io/fj8Av},
setting $\beta=1$ for deletion \citep{geva2019discofuse}.} 

\begin{table}[tb]
\centering
\footnotesize
\begin{tabular}{@{}*{4}{>{\ttfamily}c}@{}}
\toprule
\textrm{DfWiki} & \textrm{WikiSplit} & \textrm{AS} & \textrm{GEC}\\
\cmidrule(r){1-1}
\cmidrule(lr){2-2}
\cmidrule(lr){3-3}
\cmidrule(l){4-4}
, & .\textvisiblespace\ssplit{} & , & , \\
and & , & . & . \\
however\textvisiblespace\,, & .\textvisiblespace\ssplit{}\textvisiblespace he & the & the \\
,\textvisiblespace but & .\textvisiblespace\ssplit{}\textvisiblespace it & a & a \\
he & the & \& & to \\
because & and & and & in \\
,\textvisiblespace although & was & is & of \\
but & is & in & on \\
,\textvisiblespace and & '' & " & at \\
although & .\textvisiblespace\ssplit{}\textvisiblespace she & 's & for \\
his & .\textvisiblespace\ssplit{}\textvisiblespace it\textvisiblespace is & with & have \\
,\textvisiblespace while & a & for & is \\
it & .\textvisiblespace\ssplit{}\textvisiblespace they & of & was \\
,\textvisiblespace which & .\textvisiblespace\ssplit{}\textvisiblespace however & n't & and \\
she & he & an & that \\
\bottomrule
\end{tabular}
\caption{The 15 most frequently added phrases in the datasets studied in this work, in order of decreasing frequency. \ssplit{} marks a sentence boundary. ``AS''/``GEC'' is short for Abstractive Summarization/Grammatical Error Correction.}
\label{tab:phrases}
\end{table}

\begin{figure}[tb]
\includegraphics[width=\columnwidth]{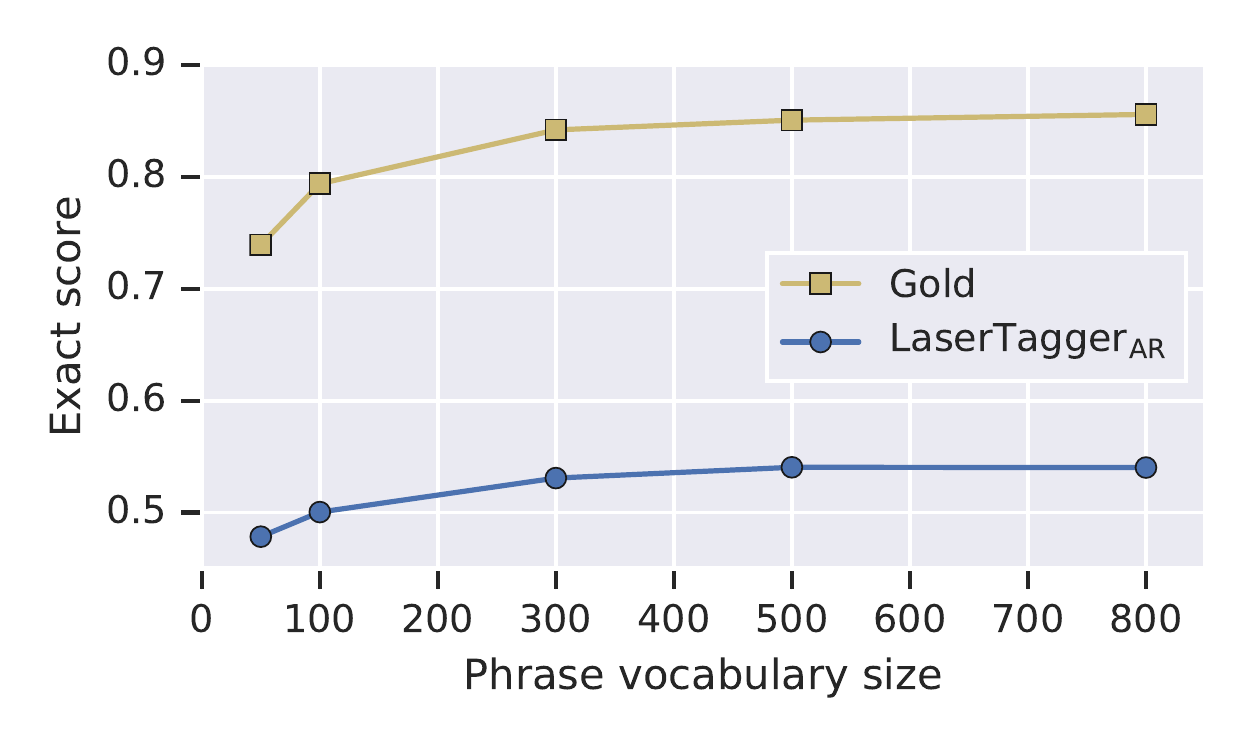}
\caption{Performance of model \taggerAR{} on the DfWiki dataset, conditioned on the vocabulary size and the gold score, i.e. the percentage of examples that can be reconstructed via text-edit operations.
}
\label{fig:vocab_size}
\end{figure}

\begin{figure*}[t]
\begin{subfigure}[t]{0.5\textwidth}
    \centering
    \includegraphics[trim=12 12 10 10,width=.9\textwidth]{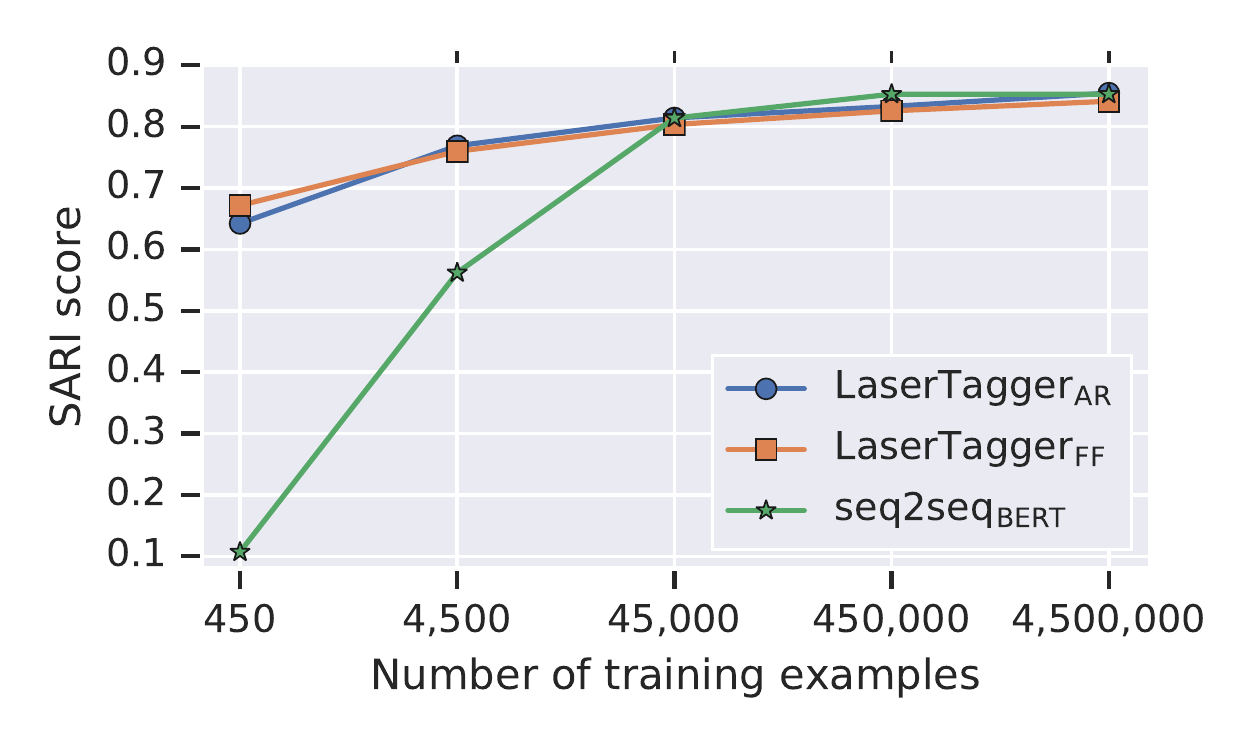}
    \caption{Sentence Fusion on DfWiki.\label{fig:data_size_df}}
\end{subfigure}\hfill
\begin{subfigure}[t]{0.5\textwidth}
    \centering
    \includegraphics[trim=12 12 10 10,width=.9\textwidth]{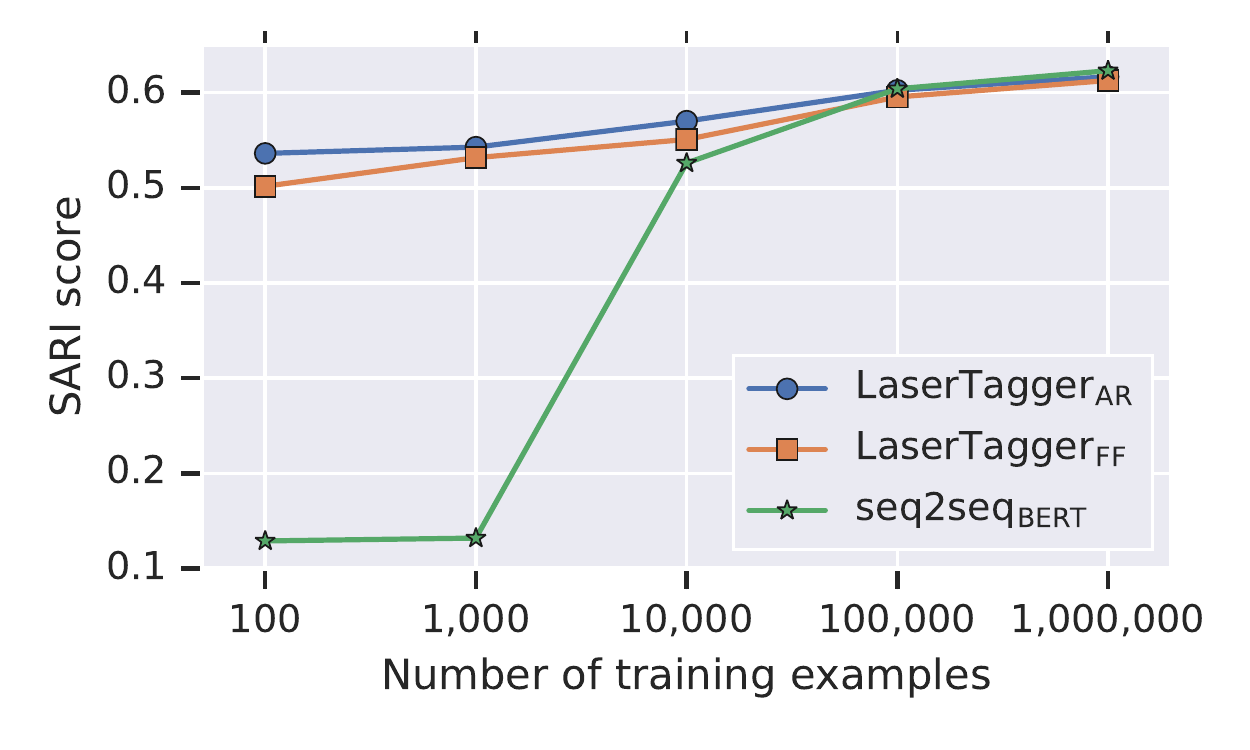}
    \caption{Split and Rephrase on WikiSplit.\label{fig:data_size_ws}}
\end{subfigure}
\caption{SARI score as a function of the training-data size for three models. Unless we have tens of thousands of training examples, the tagging approach clearly outperforms the seq2seq baseline. \label{fig:data_size}}
\end{figure*}

\textbf{Vocabulary Size.}
To understand the impact of the number of phrases we include in the vocabulary, we trained models for different vocabulary sizes (only \taggerAR). The results are shown in Figure~\ref{fig:vocab_size}. After increasing the vocabulary size to 500 phrases, Exact score reaches a plateau, so we set the vocabulary size to 500 in all the remaining experiments of this paper.\footnote{For smaller datasets, a smaller vocabulary size may yield better results, but for simplicity, we do not optimize the size separately for each dataset.} The Gold curve in Fig.~\ref{fig:vocab_size} shows that this vocabulary size is sufficient to cover 85\% of the training examples, which gives us an upper bound for the Exact score.

\textbf{Comparison against Baselines.}
Table~\ref{tab:df} lists the results for the DfWiki dataset. We obtain new SOTA results with \taggerAR, outperforming the previous SOTA 7-layer Transformer model  from \citet{geva2019discofuse} by 2.7\% Exact score and 1.0\% SARI score. We also find that the pretrained \seqtoseqbert model yields nearly as good performance, demonstrating the effectiveness of unsupervised pretraining for generation tasks. The performance of the tagger is impaired significantly when leaving out the \tag{SWAP} tag due to the model's inability to reconstruct 10.5\% of the training set.

\textbf{Impact of Dataset Size.}
We also study the effect of the training data size by creating four increasingly smaller subsets of DfWiki (see Fig.~\ref{fig:data_size_df}).\footnote{For simplicity, we use the same phrase vocabulary of size 500 computed using the validation set of 46K examples for all experiments. Note that even though some subsampled training sets contain less than 46K examples, using the same vocabulary does not give the taggers an unfair advantage over the baselines, because the tagger will never predict a phrase it has not seen in the training data.} When data size drops to 450 or 4\,500 examples, \tagger still performs surprisingly well, clearly outperforming the \seqtoseqbert baseline.

\begin{table}[tb]
\centering
\footnotesize
\begin{tabular}{lcc}
\toprule
Model & Exact & SARI \\ 
\midrule
Transformer \citep{geva2019discofuse}  & 51.1 & 84.5 \\
\seqtoseqbert & 53.6 & 85.3 \\
\midrule
\taggerAR (no \tag{SWAP}) & 46.4 & 80.4 \\
\taggerFF & 52.2 & 84.1 \\
\taggerAR & \textbf{53.8} & \textbf{85.5} \\
\bottomrule
\end{tabular}
\caption{Sentence fusion results on DfWiki.}\label{tab:df}
\end{table}

\subsection{Split and Rephrase}

The reverse task of sentence fusion is the \textit{split-and-rephrase} task, which requires rewriting a long sentence into two or more coherent short sentences.

\textbf{Data.}
We use the WikiSplit dataset~\cite{wikisplit}, which consists of 1M human-editor created examples of sentence splits, and follow the dataset split suggested by the authors. Using the phrase vocabulary of size 500 yields a 31\% coverage of the targets from the training set (top phrases shown in Table~\ref{tab:phrases}). The lower coverage compared to DfWiki suggests a higher amount of noise (due to Wikipedia-author edits unrelated to splitting).

\textbf{Results.}
\citet{wikisplit} report results using a one-layer, bi-directional LSTM (cell size 512) with attention and a copying mechanism~\cite{See2017-ue}.\footnote{\citet{wikisplit} report only BLEU but they kindly shared with us their model's predictions, allowing us to compute the Exact and SARI score for their method. Similar to their work, we used NLTK v3.2.2 for the BLEU computation.} The results are shown in Table~\ref{tab:wikisplit}. \seqtoseqbert and \taggerAR yield similar performance with each other, and they both outperform the seq2seq model with a copying mechanism from \citet{wikisplit}.

\begin{table}[tb]
\centering
\footnotesize
\begin{tabular}{lccc}
\toprule
Model & BLEU & Exact & SARI \\ 
\midrule
seq2seq \citep{wikisplit}  & 76.0 & 14.6 & 60.6 \\
\seqtoseqbert & \textbf{76.7} & 15.1 & \textbf{62.3} \\
\midrule
\taggerFF & 76.0 & 14.4 & 61.3 \\
\taggerAR & 76.3 & \textbf{15.2} & 61.7 \\
\bottomrule
\end{tabular}
\caption{Results on the WikiSplit dataset.}\label{tab:wikisplit}
\end{table}

We again studied the impact of training-data size by subsampling the training set, see Figure~\ref{fig:data_size_ws}. Similar to the previous experiment, the \tagger methods degrade more gracefully when reducing training-data size, and start to outperform the seq2seq baseline once going below circa 10k examples. The smallest training set for \taggerAR contains merely 29 examples. Remarkably, the model is still able to learn something useful that generalizes to unseen test examples, reaching a SARI score of 53.6\% and predicting 5.2\% of the targets exactly correctly.
The following is an example prediction by the model: \\
{\small
\textbf{Source:} Delhi Public Library is a national depository library in Delhi , India , it has over 35 branches across the state . \\
\textbf{Prediction:} Delhi Public Library is a national depository library in Delhi , India . \ssplit It has over 35 branches across the state . } \\
Here the model has picked the right comma to replace with a period and a sentence separator.

\subsection{Abstractive Summarization}

The task of summarization is to reduce the length of a text while preserving its meaning.

\textbf{Dataset.}
We use the dataset from \citet{Toutanova2016-rs}, which contains 6,168 short input texts (one or two sentences) and one or more human-written summaries.
The human experts were not restricted to just deleting words when generating a summary, but were allowed to also insert new words and reorder parts of the sentence, which makes this dataset particularly suited for \emph{abstractive summarization} models.

We set the size of the phrase vocabulary to 500, as for the other tasks, and extract the phrases from the training partition. With a size of 500, we are able to cover 89\% of the training data.

\begin{table}[tb]
\centering
\resizebox{\columnwidth}{!}{  
\footnotesize
\begin{tabular}{@{}lcccc@{}}
\toprule
Model & BLEU-4 & Exact & SARI & ROUGE-L \\ 
\midrule
\citet{Filippova2015-xd} & 26.7 & 0.0 & 36.2 & 70.3 \\
\citet{ClarkeL08} & 28.5 & 0.3 & 41.5 & 77.5 \\
\citet{cohn-lapata-2008} & \phantom{0}5.1 & 0.1 & 27.4 & 40.7 \\
\citet{Rush2015-qf} & 16.2 & 0.0 & 35.6 & 62.5 \\
\seqtoseqbert{} & \phantom{0}8.3 & 0.1 & 32.1 & 52.7 \\
\midrule
\taggerFF & 33.7 & 1.5 & 44.2 & 81.9 \\
\taggerAR & \textbf{35.6} & \textbf{3.8} & \textbf{44.8} & \textbf{82.8} \\
\bottomrule
\end{tabular}
}
\caption{Results on summarization.}\label{tab:summarization}
\end{table}

\textbf{Evaluation Metrics.}
In addition to the metrics from the previous sections, we report \mbox{ROUGE-L}~\cite{lin-2004-rouge}, as this is a metric that is commonly used in the summarization literature. ROUGE-L is a recall-oriented measure computed as the longest common sub-sequence between a reference summary and a candidate summary.

\textbf{Results.}
Table~\ref{tab:summarization} compares our taggers against seq2seq baselines and systems from the literature.\footnote{Results are extracted from \citet{Toutanova2016-rs}.}
\citet{Filippova2015-xd} and \citet{ClarkeL08} proposed deletion-based approaches; the former uses a seq2seq network, the latter formulates summarization as an optimization problem that is solved via integer-linear programming.
\citet{cohn-lapata-2008} proposed an early approach to abstractive summarization via a parse-tree transducer. \citet{Rush2015-qf} developed a neural seq2seq model for abstractive summarization.

In line with the results on the subsampled fusion/splitting datasets (Figure~\ref{fig:data_size}), the tagger significantly outperforms all baselines.
This shows that even though a text-editing approach is not well-suited for extreme summarization examples (a complete paraphrase with zero lexical overlap), in practice, already a limited paraphrasing capability is enough to reach good empirical performance.

Note that the low absolute values for the Exact metric are expected, since there is a very large number of acceptable summaries.

\subsection{Grammatical Error Correction (GEC)}

GEC requires systems to identify and fix grammatical errors in a given input text.

\textbf{Data.}
We use a recent benchmark from a shared task of the 2019 Building Educational Applications workshop, specifically from the Low Resource track\footnote{\scriptsize\url{www.cl.cam.ac.uk/research/nl/bea2019st/}} \citep{bryant2019bea}. The publicly available set has 4,384 ill-formed sentences together with gold error corrections, which we split 9:1 into a training and validation partition. We again create the phrase vocabulary from the 500 most frequently added phrases in the training partition, which gives us a coverage of 40\% of the training data.

\textbf{Evaluation Metrics and Results.}
We report precision and recall, and the task's main metric $F_{0.5}$, which gives more weight to the precision of the corrections than to their recall.

Table~\ref{tab:grammar-results} compares our taggers against two baselines. Again, the tagging approach clearly outperforms the BERT-based seq2seq model, here by being more than seven times as accurate in the prediction of corrections. This can be accounted to the seq2seq model's much richer generation capacity, which the model can not properly tune to the task at hand given the small amount of training data. The tagging approach on the other hand is naturally suited to this kind of problem.

We also report the best-performing method by \citet{grundkiewicz2019neural} from the shared task for informational purposes. They train a Transformer model using a dataset which is augmented by 100 million synthetic examples and 2 million Wikipedia edits, whereas we only use 4,384 sentences from the provided training dataset.

\begin{table}[tb]
\centering
\footnotesize
\begin{tabular}{lccc}
\toprule
Model & $P$ & $R$ & $F_{0.5}$ \\
\midrule
\citet{grundkiewicz2019neural} &
\textit{70.19} & \textit{47.99} & \textit{64.24} \\
\seqtoseqbert{} & \phantom{0}6.13 & 14.14 & \phantom{0}6.91 \\
\midrule
\taggerFF & 44.17 & 24.00 & 37.82 \\
\taggerAR & \textbf{47.46} & \textbf{25.58} & \textbf{40.52} \\
\bottomrule
\end{tabular}
\caption{Results on grammatical-error correction. Note that \citet{grundkiewicz2019neural} augment the training dataset of 4,384 examples by 100 million synthetic examples and 2 million Wikipedia edits.}\label{tab:grammar-results}
\end{table}

\subsection{Inference time}
Getting state-of-the-art results often requires using larger and more complex models.
When running a model in production, one cares not only about the accuracy but also the inference time. Table~\ref{tab:latency} reports latency numbers for \tagger models and our most accurate seq2seq baseline. As one can see, the \seqtoseqbert baseline is impractical to run in production even for the smallest batch size. 
On the other hand, for a batch size 8, \taggerAR is already 10x faster than comparable-in-accuracy \seqtoseqbert baseline. This difference is due to the former model using a 1-layer decoder (instead of 12 layers) and no encoder-decoder cross attention. We also tried training \seqtoseqbert with a 1-layer decoder but it performed very poorly in terms of accuracy.
Finally, \taggerFF is more than 100x faster while being only a few accuracy points below our best reported results.

\begin{table}[tb]
\centering
\resizebox{\columnwidth}{!}{  
\begin{tabular}{@{}cccc@{}}
\toprule
batch size & \taggerFF & \taggerAR & \seqtoseqbert \\ 
\midrule
1 & 13  & 535 & 1,773 \\
8 & 47 & 668 & 8,279 \\
32 & 149 & 1,273 & 27,305 \\
\bottomrule
\end{tabular}
}
\caption{Inference time (in ms) across various batch sizes on GPU (Nvidia Tesla P100) averaged across 100 runs with random inputs.}\label{tab:latency}
\end{table}

\begin{table*}[tb]
\centering
\footnotesize
\begin{tabular}{@{}lcc>{\scriptsize}m{.96\columnwidth}@{}}
\toprule
Error type & \tagger & \seqtoseqbert & \footnotesize{}Example \\ 
\midrule
Imaginary words & \textcolor{darkgreen}{not affected} & \textcolor{red}{affected} &
In:~~\,\quad\textellipsis{} Zenica (Cyrillic: ``\textbf{\foreignlanguage{russian}{Зеница}}'') is \textellipsis{} 
\newline
Out:\quad\textellipsis{} Zenica (Cyrillic: ``\textbf{gratulation\foreignlanguage{russian}{еница}}'') is \textellipsis{}
\\
Repeated phrases & \textcolor{darkgreen}{not affected} & \textcolor{red}{affected} &
In:~~\,\quad{}I'm your employee, to serve on your company. 
\newline
Out:\quad{}I'm your \textbf{company}, to serve on your \textbf{company}.
\\
Premature end-of-sentence & \textcolor{orange}{less affected} & \textcolor{red}{affected} &
In:~~\,\quad{}By the way, my favorite football team is Manchester United, they \textellipsis{} 
\newline
Out:\quad{}By the way, my favorite football team is.
\\
Hallucinations & \textcolor{orange}{less affected} & \textcolor{red}{affected} &
In:~~\,\quad{}Tobacco smokers may also experience \textellipsis{} 
\newline
Out:\quad{}\textbf{anthropology} smokers may also experience \textellipsis{} 
\\
Coreference issues & \textcolor{red}{affected} & \textcolor{red}{affected} &
In:~~\,\quad{}She is the daughter of Alistair Crane \textellipsis{} who secretly built \textellipsis{} 
\newline
Out:\quad{}She is the daughter of Alistair Crane \textellipsis{} \ssplit{} \textbf{She} secretly built \textellipsis{} 
\\
Misleading rephrasing & \textcolor{red}{affected} & \textcolor{red}{affected} &
In:~~\,\quad\textellipsis{} postal service was in no way responsible \textellipsis{} 
\newline
Out:\quad\textellipsis{} postal service \textbf{was responsible} \textellipsis{} 
\\
Lazy sentence splitting & \textcolor{red}{affected} & \textcolor{darkgreen}{not affected} &
In:~~\,\quad{}Home world of the Marglotta located in the Sagittarius Arm.
\newline
Out:\quad{}Home world of the Marglotta . \ssplit{} Located in the Sagittarius Arm.
\\
\bottomrule
\end{tabular}
\caption{Main error patterns observed in the output of the tagging and seq2seq models on their test sets (all tasks).}\label{tab:error-classes}
\end{table*}

\subsection{Qualitative evaluation}
\label{sec:qualitative-evaluation}

To assess the qualitative difference between the outputs of \tagger and \seqtoseqbert, we analyzed the texts generated by the models on the test sets of the four tasks. We inspected the respective worst predictions from each model according to BLEU and identified seven main error patterns, two of which are specific to the seq2seq model, and one being specific to \tagger.

This illustrates that \tagger is less prone to errors compared to the standard seq2seq approach, due to the restricted flexibility of its model. Certain types of errors, namely imaginary words and repeated phrases, are virtually impossible for the tagger to make. The likelihood of others, such hallucination and abrupt sentence ending, is at least greatly reduced.

In Table~\ref{tab:error-classes}, we list the error classes and refer to Appendix~\ref{app:errors} for more details on our observations.

\section{Conclusions}

We proposed a text-editing approach to text-generation tasks with high overlap between input and output texts. Compared to the seq2seq models typically applied in this setting, our approach results in a simpler sequence-tagging problem with a much smaller output tag vocabulary. We demonstrated that this approach has comparable performance when trained on medium-to-large datasets, and clearly outperforms a strong seq2seq baseline when the number of training examples is limited. Qualitative analysis of the model outputs suggests that our tagging approach is less affected by the common errors of the seq2seq models, such as hallucination and abrupt sentence ending. We further demonstrated that tagging can speed up inference by more than two orders of magnitude, making it more attractive for production applications.

\textbf{Limitations.}
Arbitrary word reordering is not feasible with our approach, although limited reordering can be achieved with deletion and insertion operations, as well as custom tags, such as \tag{SWAP} (see Section~ \ref{sec:operations}).
To enable more flexible reordering, it might be possible to apply techniques developed for phrase-based machine translation. Another limitation is that our approach may not be straightforward to apply to languages that are morphologically richer than English, where a more sophisticated realizer might be needed to adjust, e.g., the cases of the words.

In future work, we would like to experiment with more light-weight tagging architectures \citep{andor-etal-2016-globally} to better understand the trade-off between inference time and model accuracy.

\section*{Acknowledgments}

We would like to thank Enrique Alfonseca, Idan Szpektor, and Orgad Keller for useful discussions.

\balance

\bibliography{emnlp-ijcnlp-2019}

\begin{thebibliography}{49}
\expandafter\ifx\csname natexlab\endcsname\relax\def\natexlab#1{#1}\fi

\bibitem[{Andor et~al.(2016)Andor, Alberti, Weiss, Severyn, Presta, Ganchev,
  Petrov, and Collins}]{andor-etal-2016-globally}
Daniel Andor, Chris Alberti, David Weiss, Aliaksei Severyn, Alessandro Presta,
  Kuzman Ganchev, Slav Petrov, and Michael Collins. 2016.
\newblock \href {https://www.aclweb.org/anthology/P16-1231} {Globally
  normalized transition-based neural networks}.
\newblock In \emph{Proceedings of the 54th Annual Meeting of the Association
  for Computational Linguistics (Volume 1: Long Papers)}, pages 2442--2452.

\bibitem[{Botha et~al.(2018)Botha, Faruqui, Alex, Baldridge, and
  Das}]{wikisplit}
Jan~A Botha, Manaal Faruqui, John Alex, Jason Baldridge, and Dipanjan Das.
  2018.
\newblock Learning to split and rephrase from wikipedia edit history.
\newblock In \emph{Proceedings of the 2018 Conference on Empirical Methods in
  Natural Language Processing}.

\bibitem[{Bryant et~al.(2019)Bryant, Felice, Andersen, and
  Briscoe}]{bryant2019bea}
Christopher Bryant, Mariano Felice, {\O}istein~E Andersen, and Ted Briscoe.
  2019.
\newblock The bea-2019 shared task on grammatical error correction.
\newblock In \emph{Proceedings of the Fourteenth Workshop on Innovative Use of
  NLP for Building Educational Applications}, pages 52--75.

\bibitem[{Chollampatt and Ng(2018)}]{chollampatt-ng-2018}
Shamil Chollampatt and Hwee~Tou Ng. 2018.
\newblock \href {https://www.aclweb.org/anthology/D18-1274} {Neural quality
  estimation of grammatical error correction}.
\newblock In \emph{Proceedings of the 2018 Conference on Empirical Methods in
  Natural Language Processing}, pages 2528--2539.

\bibitem[{Clarke and Lapata(2008)}]{ClarkeL08}
James Clarke and Mirella Lapata. 2008.
\newblock \href {https://doi.org/10.1613/jair.2433} {Global inference for
  sentence compression: An integer linear programming approach}.
\newblock \emph{J. Artif. Intell. Res.}, 31:399--429.

\bibitem[{Cohn and Lapata(2008)}]{cohn-lapata-2008}
Trevor Cohn and Mirella Lapata. 2008.
\newblock \href {https://www.aclweb.org/anthology/C08-1018} {Sentence
  compression beyond word deletion}.
\newblock In \emph{Proceedings of the 22nd International Conference on
  Computational Linguistics (Coling 2008)}, pages 137--144.

\bibitem[{Dernoncourt et~al.(2018)Dernoncourt, Ghassemi, and
  Chang}]{Dernoncourt2018-op}
Franck Dernoncourt, Mohammad Ghassemi, and Walter Chang. 2018.
\newblock A repository of corpora for summarization.
\newblock In \emph{Proceedings of the Eleventh International Conference on
  Language Resources and Evaluation ({LREC-2018})}.

\bibitem[{Devlin et~al.(2019)Devlin, Chang, Lee, and
  Toutanova}]{devlin2018bert}
Jacob Devlin, Ming-Wei Chang, Kenton Lee, and Kristina Toutanova. 2019.
\newblock {BERT}: Pre-training of deep bidirectional transformers for language
  understanding.
\newblock In \emph{Proceedings of the 2019 Conference of the North American
  Chapter of the Association for Computational Linguistics: Human Language
  Technologies, Volume 1 (Long and Short Papers)}, pages 4171--4186.

\bibitem[{Dong et~al.(2019)Dong, Li, Rezagholizadeh, and
  Cheung}]{dong2019editnts}
Yue Dong, Zichao Li, Mehdi Rezagholizadeh, and Jackie Chi~Kit Cheung. 2019.
\newblock {EditNTS}: An neural programmer-interpreter model for sentence
  simplification through explicit editing.
\newblock In \emph{Proceedings of the 57th Annual Meeting of the Association
  for Computational Linguistics}.

\bibitem[{Filippova et~al.(2015)Filippova, Alfonseca, Colmenares, Kaiser, and
  Vinyals}]{Filippova2015-xd}
Katja Filippova, Enrique Alfonseca, Carlos~A Colmenares, Lukasz Kaiser, and
  Oriol Vinyals. 2015.
\newblock Sentence compression by deletion with {LSTMs}.
\newblock In \emph{Proceedings of the 2015 Conference on Empirical Methods in
  Natural Language Processing}, pages 360--368.

\bibitem[{Filippova and Strube(2008)}]{Filippova2008-qk}
Katja Filippova and Michael Strube. 2008.
\newblock Dependency tree based sentence compression.
\newblock In \emph{Proceedings of the Fifth International Natural Language
  Generation Conference}, pages 25--32.

\bibitem[{Ge et~al.(2018)Ge, Wei, and Zhou}]{ge-etal-2018-fluency}
Tao Ge, Furu Wei, and Ming Zhou. 2018.
\newblock \href {https://www.aclweb.org/anthology/P18-1097} {Fluency boost
  learning and inference for neural grammatical error correction}.
\newblock In \emph{Proceedings of the 56th Annual Meeting of the Association
  for Computational Linguistics (Volume 1: Long Papers)}, pages 1055--1065.

\bibitem[{Geva et~al.(2019)Geva, Malmi, Szpektor, and
  Berant}]{geva2019discofuse}
Mor Geva, Eric Malmi, Idan Szpektor, and Jonathan Berant. 2019.
\newblock {DiscoFuse}: A large-scale dataset for discourse-based sentence
  fusion.
\newblock In \emph{Proceedings of the 2019 Conference of the North American
  Chapter of the Association for Computational Linguistics: Human Language
  Technologies, Volume 1 (Long and Short Papers)}, pages 3443--3455.

\bibitem[{Grundkiewicz et~al.(2019)Grundkiewicz, Junczys-Dowmunt, and
  Heafield}]{grundkiewicz2019neural}
Roman Grundkiewicz, Marcin Junczys-Dowmunt, and Kenneth Heafield. 2019.
\newblock Neural grammatical error correction systems with unsupervised
  pre-training on synthetic data.
\newblock In \emph{Proceedings of the Fourteenth Workshop on Innovative Use of
  NLP for Building Educational Applications}, pages 252--263.

\bibitem[{Gu et~al.(2016)Gu, Lu, Li, and Li}]{gu2016incorporating}
Jiatao Gu, Zhengdong Lu, Hang Li, and Victor~OK Li. 2016.
\newblock Incorporating copying mechanism in sequence-to-sequence learning.
\newblock In \emph{Proceedings of the 54th Annual Meeting of the Association
  for Computational Linguistics (Volume 1: Long Papers)}, pages 1631--1640.

\bibitem[{Gu et~al.(2019)Gu, Wang, and Zhao}]{gu2019levenshtein}
Jiatao Gu, Changhan Wang, and Jake Zhao. 2019.
\newblock Levenshtein transformer.
\newblock \emph{arXiv preprint arXiv:1905.11006}.

\bibitem[{Jin et~al.(2019)Jin, Jin, Mueller, Matthews, and
  Santus}]{jin2019unsupervised}
Zhijing Jin, Di~Jin, Jonas Mueller, Nicholas Matthews, and Enrico Santus. 2019.
\newblock Unsupervised text style transfer via iterative matching and
  translation.
\newblock \emph{arXiv preprint arXiv:1901.11333}.

\bibitem[{Jing and McKeown(2000)}]{Jing2000-hj}
Hongyan Jing and Kathleen McKeown. 2000.
\newblock Cut and paste based text summarization.
\newblock In \emph{1st Meeting of the North American Chapter of the Association
  for Computational Linguistics}.

\bibitem[{Junczys-Dowmunt and
  Grundkiewicz(2014)}]{junczys-dowmunt-grundkiewicz-2014-amu}
Marcin Junczys-Dowmunt and Roman Grundkiewicz. 2014.
\newblock \href {https://doi.org/10.3115/v1/W14-1703} {The {AMU} system in the
  {C}o{NLL}-2014 shared task: Grammatical error correction by data-intensive
  and feature-rich statistical machine translation}.
\newblock In \emph{Proceedings of the Eighteenth Conference on Computational
  Natural Language Learning: Shared Task}, pages 25--33.

\bibitem[{Junczys-Dowmunt et~al.(2018)Junczys-Dowmunt, Grundkiewicz, Guha, and
  Heafield}]{junczys-dowmunt-etal-2018-approaching}
Marcin Junczys-Dowmunt, Roman Grundkiewicz, Shubha Guha, and Kenneth Heafield.
  2018.
\newblock \href {https://doi.org/10.18653/v1/N18-1055} {Approaching neural
  grammatical error correction as a low-resource machine translation task}.
\newblock In \emph{Proceedings of the 2018 Conference of the North {A}merican
  Chapter of the Association for Computational Linguistics: Human Language
  Technologies, Volume 1 (Long Papers)}, pages 595--606.

\bibitem[{Kasewa et~al.(2018)Kasewa, Stenetorp, and
  Riedel}]{kasewa-etal-2018-wronging}
Sudhanshu Kasewa, Pontus Stenetorp, and Sebastian Riedel. 2018.
\newblock \href {https://www.aclweb.org/anthology/D18-1541} {Wronging a right:
  Generating better errors to improve grammatical error detection}.
\newblock In \emph{Proceedings of the 2018 Conference on Empirical Methods in
  Natural Language Processing}, pages 4977--4983.

\bibitem[{Knight and Chander(1994)}]{KnightC94}
Kevin Knight and Ishwar Chander. 1994.
\newblock \href {http://www.aaai.org/Library/AAAI/1994/aaai94-119.php}
  {Automated postediting of documents}.
\newblock In \emph{Proceedings of the 12th National Conference on Artificial
  Intelligence, Seattle, WA, USA, July 31 - August 4, 1994, Volume 1.}, pages
  779--784.

\bibitem[{Lin(2004)}]{lin-2004-rouge}
Chin-Yew Lin. 2004.
\newblock \href {https://www.aclweb.org/anthology/W04-1013} {{ROUGE}: A package
  for automatic evaluation of summaries}.
\newblock In \emph{Text Summarization Branches Out: Proceedings of the {ACL}-04
  Workshop}, pages 74--81.

\bibitem[{Liu(2019)}]{Liu2019-ut}
Yang Liu. 2019.
\newblock Fine-tune bert for extractive summarization.
\newblock \emph{arXiv preprint arXiv:1903.10318}.

\bibitem[{Moryossef et~al.(2019)Moryossef, Goldberg, and
  Dagan}]{moryossef2019step}
Amit Moryossef, Yoav Goldberg, and Ido Dagan. 2019.
\newblock Step-by-step: Separating planning from realization in neural
  data-to-text generation.
\newblock In \emph{Proceedings of the 2019 Conference of the North American
  Chapter of the Association for Computational Linguistics: Human Language
  Technologies, Volume 1 (Long and Short Papers)}, pages 2267--2277.

\bibitem[{Narayan et~al.(2018)Narayan, Cohen, and Lapata}]{Narayan2018-nn}
Shashi Narayan, Shay~B Cohen, and Mirella Lapata. 2018.
\newblock Ranking sentences for extractive summarization with reinforcement
  learning.
\newblock In \emph{Proceedings of the 2018 Conference of the North American
  Chapter of the Association for Computational Linguistics: Human Language
  Technologies, Volume 1 (Long Papers)}, pages 1747--1759.

\bibitem[{Ng et~al.(2014)Ng, Wu, Briscoe, Hadiwinoto, Susanto, and
  Bryant}]{ng-etal-2014-conll}
Hwee~Tou Ng, Siew~Mei Wu, Ted Briscoe, Christian Hadiwinoto, Raymond~Hendy
  Susanto, and Christopher Bryant. 2014.
\newblock \href {https://doi.org/10.3115/v1/W14-1701} {The {C}o{NLL}-2014
  shared task on grammatical error correction}.
\newblock In \emph{Proceedings of the Eighteenth Conference on Computational
  Natural Language Learning: Shared Task}, pages 1--14.

\bibitem[{Ng et~al.(2013)Ng, Wu, Wu, Hadiwinoto, and
  Tetreault}]{ng-etal-2013-conll}
Hwee~Tou Ng, Siew~Mei Wu, Yuanbin Wu, Christian Hadiwinoto, and Joel Tetreault.
  2013.
\newblock \href {https://www.aclweb.org/anthology/W13-3601} {The {C}o{NLL}-2013
  shared task on grammatical error correction}.
\newblock In \emph{Proceedings of the Seventeenth Conference on Computational
  Natural Language Learning: Shared Task}, pages 1--12.

\bibitem[{Nikolov and Hahnloser(2018)}]{nikolov2018large}
Nikola~I Nikolov and Richard~HR Hahnloser. 2018.
\newblock Large-scale hierarchical alignment for author style transfer.
\newblock \emph{arXiv preprint arXiv:1810.08237}.

\bibitem[{Paulus et~al.(2017)Paulus, Xiong, and Socher}]{Paulus2017-dx}
Romain Paulus, Caiming Xiong, and Richard Socher. 2017.
\newblock A deep reinforced model for abstractive summarization.
\newblock \emph{arXiv preprint arXiv:1705.04304}.

\bibitem[{Prabhakaran et~al.(2018)Prabhakaran, Griffiths, Su, Verma, Morgan,
  Eberhardt, and Jurafsky}]{prabhakaran-etal-2018-detecting}
Vinodkumar Prabhakaran, Camilla Griffiths, Hang Su, Prateek Verma, Nelson
  Morgan, Jennifer~L. Eberhardt, and Dan Jurafsky. 2018.
\newblock \href {https://doi.org/10.1162/tacl_a_00031} {Detecting institutional
  dialog acts in police traffic stops}.
\newblock \emph{Transactions of the Association for Computational Linguistics},
  6:467--481.

\bibitem[{Puduppully et~al.(2018)Puduppully, Dong, and
  Lapata}]{puduppully2018data}
Ratish Puduppully, Li~Dong, and Mirella Lapata. 2018.
\newblock Data-to-text generation with content selection and planning.
\newblock \emph{arXiv preprint arXiv:1809.00582}.

\bibitem[{Rao and Tetreault(2018)}]{rao2018dear}
Sudha Rao and Joel Tetreault. 2018.
\newblock Dear sir or madam, may i introduce the {GYAFC} dataset: Corpus,
  benchmarks and metrics for formality style transfer.
\newblock In \emph{Proceedings of the 2018 Conference of the North American
  Chapter of the Association for Computational Linguistics: Human Language
  Technologies, Volume 1 (Long Papers)}, pages 129--140.

\bibitem[{Rei(2017)}]{rei-2017}
Marek Rei. 2017.
\newblock \href {https://doi.org/10.18653/v1/P17-1194} {Semi-supervised
  multitask learning for sequence labeling}.
\newblock In \emph{Proceedings of the 55th Annual Meeting of the Association
  for Computational Linguistics (Volume 1: Long Papers)}, pages 2121--2130.

\bibitem[{Rei et~al.(2017)Rei, Felice, Yuan, and Briscoe}]{rei-etal-2017}
Marek Rei, Mariano Felice, Zheng Yuan, and Ted Briscoe. 2017.
\newblock \href {https://doi.org/10.18653/v1/W17-5032} {Artificial error
  generation with machine translation and syntactic patterns}.
\newblock In \emph{Proceedings of the 12th Workshop on Innovative Use of {NLP}
  for Building Educational Applications}, pages 287--292.

\bibitem[{Rozovskaya et~al.(2014)Rozovskaya, Chang, Sammons, Roth, and
  Habash}]{rozovskaya-etal-2014-illinois}
Alla Rozovskaya, Kai-Wei Chang, Mark Sammons, Dan Roth, and Nizar Habash. 2014.
\newblock \href {https://doi.org/10.3115/v1/W14-1704} {The
  {I}llinois-{C}olumbia system in the {C}o{NLL}-2014 shared task}.
\newblock In \emph{Proceedings of the Eighteenth Conference on Computational
  Natural Language Learning: Shared Task}, pages 34--42.

\bibitem[{Rush et~al.(2015)Rush, Chopra, and Weston}]{Rush2015-qf}
Alexander~M Rush, Sumit Chopra, and Jason Weston. 2015.
\newblock A neural attention model for abstractive sentence summarization.
\newblock In \emph{Proceedings of the 2015 Conference on Empirical Methods in
  Natural Language Processing}, pages 379--389.

\bibitem[{See et~al.(2017)See, Liu, and Manning}]{See2017-ue}
Abigail See, Peter~J Liu, and Christopher~D Manning. 2017.
\newblock Get to the point: Summarization with {Pointer-Generator} networks.
\newblock In \emph{Proceedings of the 55th Annual Meeting of the Association
  for Computational Linguistics (Volume 1: Long Papers)}, pages 1073--1083.

\bibitem[{Sutskever et~al.(2014)Sutskever, Vinyals, and Le}]{sutskever}
Ilya Sutskever, Oriol Vinyals, and Quoc~V Le. 2014.
\newblock Sequence to sequence learning with neural networks.
\newblock In \emph{Advances in Neural Information Processing Systems 27}, pages
  3104--3112.

\bibitem[{Tan et~al.(2017)Tan, Wan, and Xiao}]{Tan2017-tt}
Jiwei Tan, Xiaojun Wan, and Jianguo Xiao. 2017.
\newblock Abstractive document summarization with a {Graph-Based} attentional
  neural model.
\newblock In \emph{Proceedings of the 55th Annual Meeting of the Association
  for Computational Linguistics (Volume 1: Long Papers)}, pages 1171--1181.

\bibitem[{Toutanova et~al.(2016)Toutanova, Brockett, Tran, and
  Amershi}]{Toutanova2016-rs}
Kristina Toutanova, Chris Brockett, Ke~M Tran, and Saleema Amershi. 2016.
\newblock A dataset and evaluation metrics for abstractive compression of
  sentences and short paragraphs.
\newblock In \emph{Proceedings of the 2016 Conference on Empirical Methods in
  Natural Language Processing}, pages 340--350.

\bibitem[{Vinterbo(2002)}]{vinterbo2002note}
Staal~A Vinterbo. 2002.
\newblock A note on the hardness of the k-ambiguity problem.
\newblock \emph{Technical Report DSG-T R-2002-006}.

\bibitem[{Wiseman et~al.(2018)Wiseman, Shieber, and Rush}]{wiseman2018learning}
Sam Wiseman, Stuart~M Shieber, and Alexander~M Rush. 2018.
\newblock Learning neural templates for text generation.
\newblock \emph{arXiv preprint arXiv:1808.10122}.

\bibitem[{Xu et~al.(2015)Xu, Ba, Kiros, Cho, Courville, Salakhudinov, Zemel,
  and Bengio}]{captioning}
Kelvin Xu, Jimmy Ba, Ryan Kiros, Kyunghyun Cho, Aaron Courville, Ruslan
  Salakhudinov, Rich Zemel, and Yoshua Bengio. 2015.
\newblock Show, attend and tell: Neural image caption generation with visual
  attention.
\newblock In \emph{Proceedings of the 32nd International Conference on Machine
  Learning}, volume~37 of \emph{Proceedings of Machine Learning Research},
  pages 2048--2057.

\bibitem[{Xu et~al.(2016{\natexlab{a}})Xu, Napoles, Pavlick, Chen, and
  Callison-Burch}]{Xu2016-mu}
Wei Xu, Courtney Napoles, Ellie Pavlick, Quanze Chen, and Chris Callison-Burch.
  2016{\natexlab{a}}.
\newblock Optimizing statistical machine translation for text simplification.
\newblock \emph{Transactions of the Association for Computational Linguistics},
  4:401--415.

\bibitem[{Xu et~al.(2016{\natexlab{b}})Xu, Napoles, Pavlick, Chen, and
  Callison-Burch}]{xu2016optimizing}
Wei Xu, Courtney Napoles, Ellie Pavlick, Quanze Chen, and Chris Callison-Burch.
  2016{\natexlab{b}}.
\newblock Optimizing statistical machine translation for text simplification.
\newblock \emph{Transactions of the Association for Computational Linguistics},
  4:401--415.

\bibitem[{Zhang and Lapata(2017)}]{Zhang2017-th}
Xingxing Zhang and Mirella Lapata. 2017.
\newblock Sentence simplification with deep reinforcement learning.
\newblock In \emph{Proceedings of the 2017 Conference on Empirical Methods in
  Natural Language Processing}, pages 584--594.

\bibitem[{Zhao et~al.(2019)Zhao, Wang, Shen, Jia, and Liu}]{zhaoAbs1903}
Wei Zhao, Liang Wang, Kewei Shen, Ruoyu Jia, and Jingming Liu. 2019.
\newblock Improving grammatical error correction via pre-training a
  copy-augmented architecture with unlabeled data.
\newblock In \emph{Proceedings of the 2019 Conference of the North American
  Chapter of the Association for Computational Linguistics: Human Language
  Technologies, Volume 1 (Long and Short Papers)}, pages 156--165.

\bibitem[{Zhu et~al.(2010)Zhu, Bernhard, and Gurevych}]{Zhu2010-mv}
Zhemin Zhu, Delphine Bernhard, and Iryna Gurevych. 2010.
\newblock A monolingual tree-based translation model for sentence
  simplification.
\newblock In \emph{Proceedings of the 23rd International Conference on
  Computational Linguistics (Coling 2010)}, pages 1353--1361.

\end{thebibliography}
\bibliographystyle{acl_natbib}

\balance

\clearpage

\appendix \label{appendix}

\section*{Appendix}

\balance

\section{Examples from Qualitative Analysis} \label{app:errors}

In Table~\ref{tab:errors}, we present cherry-picked examples for the error patterns introduced in Section~\ref{sec:qualitative-evaluation}.

\paragraph{Imaginary words.}
Operating on a sub-word level, the seq2seq model is capable of producing nonexistent words by concatenating unrelated WordPieces. Indeed, we have encountered examples of such made-up words in the outputs of the seq2seq model on all 4 tasks. Typically, this happens when the model encounters a rare word in the input text. \tagger, which is trained on the word level, is immune to this specific type of error.

\paragraph{Premature end-of-sentence.}
A seq2seq model produces sequences of arbitrary length by generating the end-of-sentence (\texttt{EOS}). We have seen the seq2seq model generate \texttt{EOS} prematurely, resulting in an abrupt sentence ending. In extreme cases, and especially on the abstractive summarization task, the model generated \texttt{EOS} at the start, effectively producing an empty output. For \tagger, this type of error is technically possible but very unlikely, and we have not seen it in practice. (The tagger would need to generate a long sequence of \texttt{DELETE} tags---something it has not seen in the training data.)

\paragraph{Repeated phrases.}
Another type of error specific to the seq2seq model is repetition of information---either single words or entire phrases. In the sentence splitting task, the seq2seq model would often repeat parts of the sentence twice (before and after the splitting symbol). In the grammar correction and summarization tasks, the seq2seq model would often replace a rare word with another word from the input sentence, thus repeating that word twice. \tagger can only add words or phrases from its limited vocabulary, which is unlikely to cause repetition.

We observed that the seq2seq model was especially likely to repeat large fragments in the sentence splitting task in cases when there is no obvious good way to split the sentence. Interestingly, in most of these cases \tagger did not split the sentence at all, by not inserting the sentence-splitting symbol \ssplit{}, even though such examples were not present in its training data. In other cases, \tagger produced a ''lazy split'' (discussed below).

\paragraph{Hallucination} is a known problem for neural networks, but \tagger is susceptible to it to a much lesser degree. \tagger can ''hallucinate'' only by inserting an unexpected word or a short phrase from its vocabulary. We have seen such insertions in the tagger output that rendered the sentence ungrammatical, or simply odd.
A seq2seq model is more likely to produce subtly misleading hallucinations that misrepresent the input text, while looking fluent and credible.
We have seen examples of the seq2seq model changing the factual details, which is a more dangerous error to make in some scenarios.

\paragraph{Coreference problems.}
This type of errors is often made by both seq2seq and tagger models. In the most typical instance, a model inserts an incorrect pronoun. In other cases,  the model makes an incorrect coreference resolution, by inserting the wrong proper noun. The seq2seq model is more susceptible to the second type of error because of its ability to copy proper nouns from the input sentence. \tagger will typically not attempt to resolve the coreference and just leave the pronoun intact.

\paragraph{Misleading rephrasing through deletion.}
We want to highlight that, although \tagger has a limited set of operations and cannot insert arbitrary phrases, it is not immune to semantic errors and misrepresenting the input text. We have seen examples where deletion of word spans was sufficient to completely alter the meaning of the sentence. A seq2seq model is naturally also prone to this type of errors.

\paragraph{''Lazy splitting.''}
This type of error is specific to \tagger and the sentence splitting task. It occurs when the input sentence is split arbitrarily in the middle, without any modifications made to the resulting two parts. Such splits were usually made at a reasonable point in the input sentence, i.e. yielding two valid grammatical clauses, at least one of which, however, was not a complete sentence.

\begin{table*}[t]
\small
\centering
\begin{tabular}{p{0.05\textwidth}p{0.95\textwidth}}
\multicolumn{2}{l}{\textcolor{black}{Imaginary words introduced by a seq2model (sentence splitting task).}} \\
\textcolor{gray}{input:} & Zenica ( Cyrillic : '' \textbf{\textcolor{black}{\foreignlanguage{russian}{Зеница}}} '' ) is an industrial city ( the third largest , after Sarajevo and Banja Luka ) in Bosnia and Herzegovina and the capital of the Zenica - Doboj Canton of the Federation of Bosnia and Herzegovina entity . \\
\textcolor{gray}{output:} & Zenica ( Cyrillic : ''\textbf{\textcolor{black}{gratulation\foreignlanguage{russian}{еница}}} '' ) is an industrial city ( the third largest , after Sarajevo and Banja Luka ) in Bosnia and Herzegovina . \ssplit It is the capital of the Zenica - Doboj Canton of the Federation of Bosnia and Herzegovina entity . \\
\midrule
\multicolumn{2}{l}{\textcolor{black}{Imaginary words indroduced by a seq2model (sentence fusion task).}} \\
\textcolor{gray}{input:} & Carboxysomes are found in lithoautotrophically and mixotrophically grown cells . Carboxysomes aid carbon fixation . \\
\textcolor{gray}{output:} & Carboxysomes , which aid carbon \textbf{fixotrophically} and mixotrophically grown cells . \\
\midrule
\multicolumn{2}{l}{\textcolor{black}{Abrupt sentence end by the seq2seq model (grammar correction task).}} \\
\textcolor{gray}{input:} & By the way, my favorite football team is Manchester United, they are brilliant, they have an amazing football players, and they are awesome. \\
\textcolor{gray}{output:} & By the way, my favorite football team is. \\
\midrule
\multicolumn{2}{l}{\textcolor{black}{Abrupt sentence end by the seq2seq model (summarization task).}} \\
\textcolor{gray}{input:} & It is, I suppose, likely that people with little money must think of money more than the well-off do and that the hungry will dwell on thoughts of food and the thirsty on drink. \\
\textcolor{gray}{output:} & \textbf{It'} \\
\midrule
\multicolumn{2}{l}{\textcolor{black}{Redundant repetition by the seq2seq model (sentence splitting task).}} \\
\textcolor{gray}{input:} &  Bethany Noel Mota ( born November 7 , 1995 ) is an American video blogger from Los Banos , California . \\
\textcolor{gray}{output:} & Bethany Noel \textbf{Mota ( born November 7 , 1995 ) is an American video blogger from Los Banos , California .} \ssplit{} \textbf{Mota is an American video blogger from Los Banos , California .} \\
\midrule
\multicolumn{2}{l}{\textcolor{black}{Redundant repetition by the seq2seq model (sentence fusion task).}} \\
\textcolor{gray}{input:} &  General Fazlollah Zahedi became prime minister . General Fazlollah Zahedi led the military coup . \\
\textcolor{gray}{output:} & General Fazlollah Zahedi , who led the military coup , \textbf{became prime minister} and \textbf{became prime minister} . \\
\midrule
\multicolumn{2}{l}{\textcolor{black}{Redundant repetition by the seq2seq model (grammar correction task).}} \\
\textcolor{gray}{input:} &  I'm your employee, to serve on your company. \\
\textcolor{gray}{output:} &  I'm your \textbf{company}, to serve on your \textbf{company}. . \\
\midrule
\multicolumn{2}{l}{\textcolor{black}{Hallucinations produced by the tagger (grammar correction task).}} \\
\textcolor{gray}{input:} &  If you are interesedt, find me at 8 clock near the cinema entrance. \\
\textcolor{gray}{output:} &  If you are \textbf{play}, find me at 8 clock near the cinema entrance. \\
\midrule
\multicolumn{2}{l}{\textcolor{black}{Hallucinations produced by the tagger (sentence fusion task).}} \\
\textcolor{gray}{input:} &  He carried a bow . He was Nerva . \\
\textcolor{gray}{output:} &  \textbf{Receiving} a bow , he was Nerva . \\
\midrule
\multicolumn{2}{l}{\textcolor{black}{Hallucinations produced by the seq2seq model (sentence splitting task).}} \\
\textcolor{gray}{input:} &  The native block size is 8192 bytes ; inodes are \textbf{512} bytes on disk . \\
\textcolor{gray}{output:} &  The native block size is 8192 bytes . \ssplit Inodes are \textbf{256} bytes on disk . \\
\midrule
\multicolumn{2}{l}{\textcolor{black}{Hallucinations produced by the seq2seq model (sentence fusion task).}} \\
\textcolor{gray}{input:} &  Tobacco smokers may also experience more powerful effects with psilocybin . Tobacco smoke exposure decreases the activity of MAO in the brain and peripheral organs . \\
\textcolor{gray}{output:} &  \textbf{anthropology} smokers may also experience more powerful effects with psilocybin . \\
\midrule
\multicolumn{2}{l}{\textcolor{black}{Coreference error made by the seq2seq model (sentence fusion task).}} \\
\textcolor{gray}{input:} &  He tries to help Yuushi with catching Sayuki 's eye , and encourages the idea of a baseball club . \\
\textcolor{gray}{output:} &  He tries to help Yuushi with catching Sayuki 's eye . \ssplit{} \textbf{Yuushi} encourages the idea of a baseball club . \\
\midrule
\multicolumn{2}{l}{\textcolor{black}{Pronominalization error made by the tagger (sentence fusion task).}} \\
\textcolor{gray}{input:} &  Sharmistha Mukherjee is an Indian Kathak dancer , choreographer and politician of the Indian National Congress . \\
\textcolor{gray}{output:} &  Sharmistha Mukherjee is an Indian Kathak dancer , choreographer . \ssplit{} And politician . \ssplit{} \textbf{He} is of the Indian National Congress . \\
\midrule
\multicolumn{2}{l}{\textcolor{black}{''Lazy splitting'' made by the tagger (sentence splitting task).}} \\
\textcolor{gray}{input:} &  Home world of the Marglotta located in the Sagittarius Arm . \\
\textcolor{gray}{output:} &  Home world of the Marglotta . \ssplit{} Located in the Sagittarius Arm . \\
\midrule
\multicolumn{2}{l}{\textcolor{black}{''Lazy splitting'' made by the tagger (sentence splitting task).}} \\
\textcolor{gray}{input:} &  Jesse has a son named Sahvere and a duaghter named Jaylen . \\
\textcolor{gray}{output:} &  Jesse has a son named Sahvere . \ssplit{} And a duaghter named Jaylen . \\
\midrule
\multicolumn{2}{l}{\textcolor{black}{Misleading rephrasing through deletion, made by the tagger (sentence splitting task).}} \\
\textcolor{gray}{input:} &  She has an ex-lover , Winnie Mann , with whom she has a son , Wilson and an adopted a Chinese girl , Jang Yin . \\
\textcolor{gray}{output:} &  She has an ex-lover , Winnie Mann , a son , Wilson . \ssplit{} \textbf{She was} adopted Chinese girl , Jang Yin . \\
\midrule
\multicolumn{2}{l}{\textcolor{black}{Misleading rephrasing through deletion, made by the tagger (sentence splitting task).}} \\
\textcolor{gray}{input:} &  The article proudly notes that the postal service \textbf{was in no way responsible} for the 1996 crash of Valujet Flight 592 in the Florida Everglades. \\
\textcolor{gray}{output:} &  The article notes postal service \textbf{was responsible} for the 1996 crash of Valujet Flight 592. \\
\end{tabular}
\caption{Illustration of the typical errors produced by the models (cherry-picked examples).}\label{tab:errors}
\end{table*}

\end{document}